\pdfoutput=1
\documentclass[11pt,letterpaper]{article}
\usepackage[T1]{fontenc}
\usepackage[utf8]{inputenc}
\usepackage{lmodern}
\usepackage[margin=0.85in]{geometry}
\usepackage{amsmath,amssymb}
\usepackage{graphicx}
\usepackage{booktabs,array,tabularx}
\usepackage{caption}
\usepackage{microtype}
\usepackage{enumitem}
\usepackage[section]{placeins}
\usepackage{xurl}
\usepackage[hidelinks]{hyperref}
\setlist[enumerate]{leftmargin=*,itemsep=0.3em,topsep=0.4em}

\newcolumntype{Y}{>{\raggedright\arraybackslash}X}
\makeatletter
\renewenvironment{thebibliography}[1]
 {\section*{\refname}\small\raggedright
  \list{\@biblabel{\@arabic\c@enumiv}}
       {\setlength{\leftmargin}{1.27cm}\setlength{\labelwidth}{1.05cm}
        \setlength{\labelsep}{0.22cm}\setlength{\itemindent}{0pt}
        \setlength{\itemsep}{0.5\baselineskip}\setlength{\parsep}{0pt}
        \usecounter{enumiv}\let\p@enumiv\@empty
        \renewcommand{\theenumiv}{\@arabic\c@enumiv}
        }
  \sloppy\clubpenalty4000\widowpenalty4000}
 {\endlist}
\makeatother
\title{A Three-Dimensional Reverse-Projection Method for Sparse Point Cloud Completion and Its Application to High-Speed Train Nose Reconstruction}

\author{
Xiaozhen Ma,
Zhao Tang\thanks{Corresponding author.},
Hanbin Lai,
Ruiqi Chen,
Jin Jin
}

\date{}

\hypersetup{
    pdftitle={A Three-Dimensional Reverse-Projection Method for Sparse Point Cloud Completion and Its Application to High-Speed Train Nose Reconstruction},
    pdfauthor={Xiaozhen Ma, Zhao Tang, Hanbin Lai, Ruiqi Chen, Jin Jin}
}

\begin{document}

\maketitle

\begin{abstract}
Holes in LiDAR scans of environments with glass windows remain an unresolved problem in three-dimensional reconstruction. This study presents a pipeline for point cloud acquisition, filtering, completion, and surface reconstruction to address sparse sampling and missing window regions in scans of a high-speed train nose. FAST-LIVO2 provides the initial point cloud through multisensor odometry and mapping, and moving least squares (MLS) smooths the observations. We then introduce three-axis projection-based subdivision and interpolation with reverse hole boundary identification, referred to as three-axis reverse completion. The method interpolates missing regions from observations around each hole. Greedy projection triangulation, Poisson surface reconstruction, and a Marching Cubes-based pipeline generate meshes from the completed point cloud. Experiments on a proportionally scaled display model of a high-speed train nose show that the proposed method fills missing point cloud regions around the glass windows. Under the evaluation setting used in this study, greedy projection triangulation yields lower geometric distance errors than the other two reconstruction pipelines. The pipeline supports non-contact digital modeling of train nose geometry and provides a practical approach to reconstructing objects with glass windows.
\end{abstract}


\section{Introduction}

Mobile scans of a high-speed train nose often capture dense points on the streamlined shell but leave gaps around the cab windows. The point cloud acquired in this study contains approximately 9.7 million points, yet local holes remain. The total point count therefore does not indicate whether the surface is fully observed. Adding samples near existing surfaces alone does not cover the interiors of these holes and may incorrectly connect separate surfaces. Completion requires identifying the missing regions and using the surrounding observations to constrain the positions of new points, providing a geometric basis for digital modeling and subsequent design analysis \cite{ref1,ref2,ref3}.

Multisensor simultaneous localization and mapping (SLAM) methods fuse sensor measurements for pose estimation and point cloud registration \cite{ref4,ref5}. Local surface fitting and point cloud upsampling improve the smoothness and sampling density of existing observations \cite{ref6,ref7,ref8}. However, missing regions present two specific difficulties. The absence of observations inside a hole makes its extent difficult to determine directly, while nearby points may belong to different sides of the object, causing interpolation to mix geometry from separate surfaces. Learning-based completion recovers missing structures by learning the relationship between partial observations and complete shapes \cite{ref9}. Examples include PointAttN, which uses attention \cite{ref10}, and SuperPC, which uses a unified diffusion model \cite{ref11}. This study instead uses the geometry of the target point cloud to locate local holes and constrain completion with observations from the same surface layer in multiple directions, without relying on a pretrained shape model.

To address these difficulties, we propose three-axis projection-based subdivision and interpolation with reverse hole boundary identification, hereafter referred to as three-axis reverse completion. Projections along three orthogonal directions reduce the dependence of hole identification on a single viewing direction. Unoccupied locations in each projection domain become candidate hole points, allowing regions without observations to participate in clustering. The method then separates overlapping surface layers, retrieves observations from the same layer in multiple directions around each hole, and applies inverse distance weighting to reduce cross-layer mixing and one-sided neighborhood support. Combined with FAST-LIVO2 acquisition, MLS preprocessing, and surface reconstruction, the method forms the train nose reconstruction pipeline shown in Figure 1.

The main contributions of this study are as follows:

\begin{enumerate}

\item We propose three-axis reverse completion. The method converts unoccupied grid locations within valid projection domains into candidate hole points, identifies missing regions through clustering, and establishes their correspondence with surrounding observed boundaries to support the completion of regional holes.

\item We develop a multidirectional interpolation strategy using observations from the same surface layer. To address overlapping projections and uneven sampling along hole boundaries, the strategy combines surface layer separation with multidirectional boundary searches to constrain new points to the appropriate layer. The completed points from each projection direction are then recovered in three-dimensional space.

\item We validate the method on a point cloud of a proportionally scaled train nose display model acquired by our research team. We integrate the method into a multisensor point cloud reconstruction pipeline, compare it with nearest-neighbor interpolation, and compare the results of greedy projection triangulation, Poisson surface reconstruction, and Marching Cubes to examine window completion and differences in surface reconstruction.

\end{enumerate}

\begin{figure}[htbp]
\centering
\includegraphics[width=\linewidth]{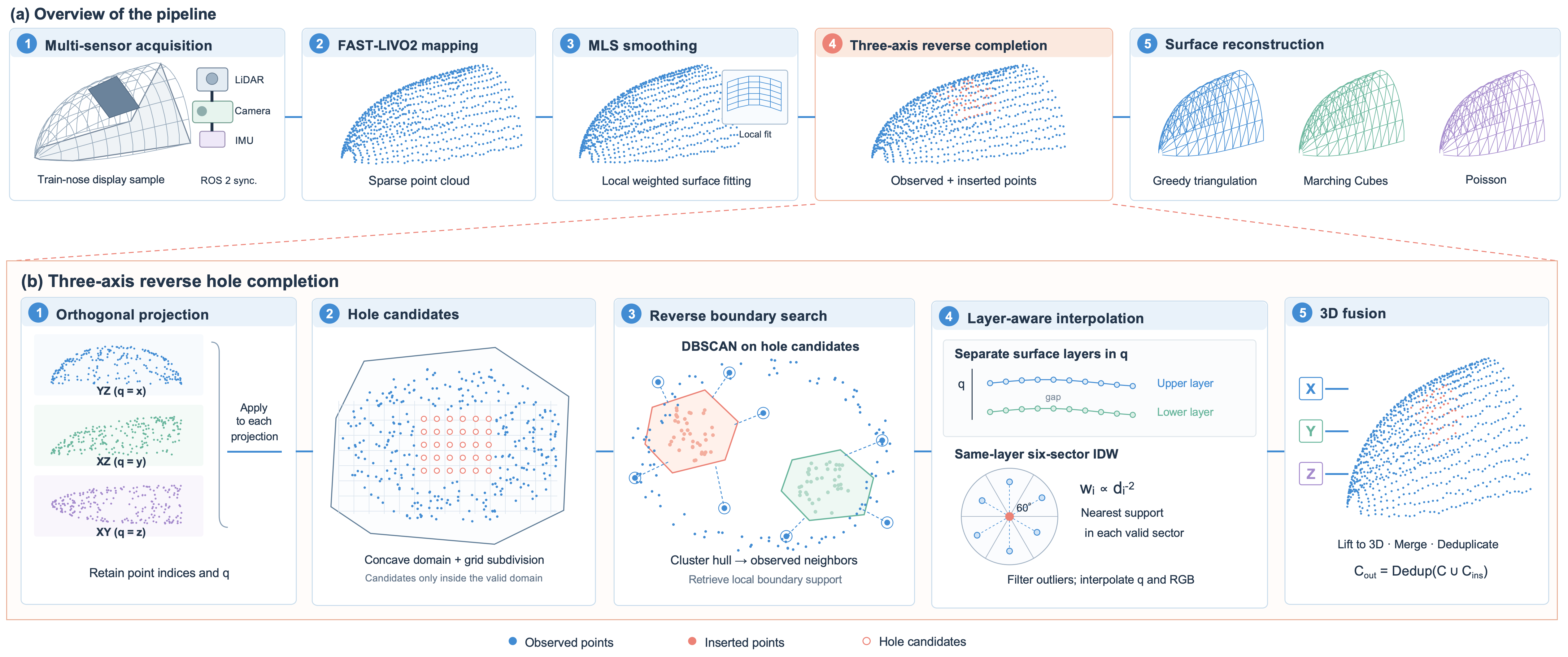}
\caption{Overall pipeline for three-axis reverse completion and train nose reconstruction.}
\label{fig:1}
\end{figure}

\section{Related Work}

Reconstructing a train nose from point clouds involves three connected stages: processing the observations, completing missing regions, and reconstructing the surface. Mobile scans require registration in a common coordinate system and filtering to reduce measurement noise. Densification or completion then addresses the sampling density of observed surfaces and the distribution of missing regions before surface reconstruction converts the discrete points into a continuous representation. The following discussion reviews work relevant to these stages.

\subsection{Geometry-Based Point Cloud Completion}

Sparse point cloud SLAM has advanced significantly in recent years, but its applications primarily involve robot localization and mapping in settings that are insensitive to holes in the point cloud. Geometric hole filling either operates directly on the point cloud or first constructs a mesh \cite{ref12}. Mesh-based methods use boundary connectivity to generate filling patches. For example, Gou et al. introduce LIMOFilling, which detects boundary holes through local overlap analysis and uses neighboring structures to recover missing sharp features \cite{ref13}. Hole detection also requires distinguishing the outer boundary of a model from internal hole boundaries. Yip et al. address this problem in edge-manifold triangular meshes with singular vertices through boundary traversal, complex boundary decomposition, and hole classification \cite{ref14}. For gaps in terrestrial LiDAR data, Kang et al. compare nearest-neighbor, inverse distance, triangulated irregular network, spline, and Kriging interpolation \cite{ref15}. Unlike deep learning-based completion, these geometric approaches generate new points from the boundaries and local surface constraints of the target data. Lan et al. combine boundary extraction, iterative slicing, and MLS with additional increment conditions for hole filling \cite{ref16}. Zhang et al. identify holes using density and boundary information in two-dimensional projections and repair them using geometry and curvature \cite{ref17}. These studies show that constructing candidate hole regions and selecting neighborhood constraints are central to methods that do not require training. For multiple holes, Hou et al. identify boundaries using a fitted global plane and fill the holes through local surface fitting near those boundaries \cite{ref18}. Xie et al. use signal decomposition to formulate terrain hole filling with ambiguous boundaries as B-spline surface fitting and two-dimensional image inpainting \cite{ref19}. For reflective regions, Sun et al. combine outlier identification, hole filling, and Poisson surface reconstruction to address missing observations \cite{ref20}. Together, these studies motivate the joint treatment of missing-region identification and local geometric constraints. Our work focuses on hole localization and support point selection in sparse point clouds.

\subsection{Deep Learning-Based Point Cloud Completion}

Machine learning methods typically map sparse point sets to denser ones using point features and neighborhood relationships. PU-Net expands point features, while PU-GCN uses graph convolutions \cite{ref7,ref8}. PU-GAN adds adversarial learning to constrain the distribution of generated points \cite{ref21}. The underlying representation also affects point generation. PU-VoxelNet combines voxel representations with density-guided resampling for arbitrary-scale upsampling \cite{ref22}, while PU-Gaussian samples and refines points using local anisotropic Gaussian representations \cite{ref23}. These methods primarily improve sampling density and point distribution on observed surfaces. Recovering extended missing regions still requires additional geometric information.

For extended missing regions, learning-based completion typically learns the relationship between partial observations and complete structures from training data \cite{ref9}. PointAttN uses attention to model relationships between points \cite{ref10}, while SuperPC unifies completion, upsampling, denoising, and colorization within a diffusion model \cite{ref11}.

For the window regions of a train nose, our method constructs candidate missing regions from unoccupied locations in three orthogonal projections. It clusters the candidate hole points to localize individual holes and applies inverse distance weighted interpolation with multidirectional support from the same surface layer. The distinction lies in how hole localization and support point selection work together, rather than in the use of projection or interpolation alone. Completion relies on observations around each hole and does not require a pretrained shape model.

\subsection{From Discrete Points to Continuous Surfaces}

Geometric surface reconstruction establishes connections between discrete points through local meshing or global surface fitting. Greedy projection triangulation creates triangular faces from neighborhood projections \cite{ref24}. Poisson surface reconstruction fits an implicit function to recover a global surface, while screened Poisson reconstruction adds point position constraints to keep the surface closer to the observations \cite{ref25,ref26}. For pipelines that use voxel-based scalar fields, Marching Cubes generates a mesh by extracting an isosurface \cite{ref27}. Although these methods address connectivity and smoothness in different ways, reconstruction across extended gaps still depends on the geometry available around them.

Neural implicit representations and three-dimensional Gaussian representations also support scene reconstruction. Huang et al. introduce neural kernel surface reconstruction to estimate implicit surfaces from sparse and noisy point clouds \cite{ref28}. Shi et al. combine local signed distance functions into a lightweight neural representation for large-scale three-dimensional mapping \cite{ref29}. With a different emphasis from direct geometric surface recovery, GS-LIVM fuses LiDAR, camera, and IMU observations and represents scenes with three-dimensional Gaussians for real-time photorealistic mapping \cite{ref30}. These approaches address continuous geometry and scene appearance, respectively. Our study focuses on completing missing point cloud regions for train nose surface reconstruction and evaluates the resulting meshes through their shape and geometric distance errors.

These studies offer several representations for recovering continuous surfaces from discrete observations. For large missing regions such as train windows, however, surface reconstruction first requires sufficient geometric samples. We address this requirement at the point cloud level through three-axis reverse completion, then compare different reconstruction pipelines in terms of the shape and geometric distance errors of the resulting surfaces.

\section{Method}

The proposed pipeline reconstructs the surface of a high-speed train nose from sparse point clouds. It combines multisensor point cloud acquisition with FAST-LIVO2, MLS filtering with normal estimation, three-axis reverse completion, and a comparison of surface reconstruction algorithms.

\subsection{Multisensor Point Cloud Acquisition}

\subsubsection{Acquisition System}

We develop a portable three-dimensional reconstruction device that combines a camera, a light detection and ranging (LiDAR) sensor, and an inertial measurement unit (IMU). LiDAR provides global geometric observations, the camera supplies visual features and texture, and the IMU records motion. These complementary measurements support handheld mobile scanning of a proportionally scaled display model of a high-speed train nose and provide the initial point cloud for subsequent reconstruction.

\subsubsection{Point Cloud Acquisition with FAST-LIVO2}

The acquisition system uses the open-source FAST-LIVO2 framework from the MaRS Laboratory at the University of Hong Kong \cite{ref5}, with Robot Operating System 2 (ROS 2) software adaptation and time synchronization for the device \cite{ref31}.

FAST-LIVO2 uses an error-state iterated Kalman filter (ESIKF) to estimate the system state from LiDAR, camera, and IMU measurements through sequential LiDAR-inertial and visual-inertial updates. Scan reprojection handles sensor synchronization, while forward and backward propagation correct motion distortion. The LiDAR module registers raw points directly to a voxel map and incorporates beam divergence into its noise model. The visual module uses LiDAR plane priors for affine warping, updates the state by minimizing photometric error, and estimates exposure time online to account for illumination changes. A unified voxel map supports visible-voxel queries, on-demand ray casting, and outlier rejection to select valid visual points, followed by normal refinement and reference patch updates. A ring buffer supports efficient local map maintenance \cite{ref5}.

\subsection{MLS Filtering and Interpolation by Weighted Residual Minimization}

MLS smooths and interpolates a point cloud by fitting a local surface around each query point. The fitting process minimizes the weighted sum of squared residuals \cite{ref6,ref32}.

\subsubsection{Local Fitting Function}

For a query point x, a k-d tree search identifies the neighboring samples $\{x_i\}_{i=1}^{n}$. We seek a fitting function f in the polynomial space $\Pi_m$ of degree m by minimizing

\begin{equation}
\min_{f\in\Pi_m}\sum_{i=1}^{n}\omega(\lVert x-x_i\rVert)\left[f(x_i)-y_i\right]^2.
\label{eq:1}
\end{equation}

Here, $y_i$ is the observed value at a neighboring sample, and $\omega$ is a Gaussian weighting function:

\begin{equation}
\omega(d)=\exp(-d^2/h^2),\qquad d=\lVert x-x_i\rVert.
\label{eq:2}
\end{equation}

The parameter h controls the influence radius.

\subsubsection{Polynomial Basis}

The fitting function is a linear combination of polynomial basis functions:

\begin{equation}
f(x)=p(x)^{\mathsf T}\alpha.
\label{eq:3}
\end{equation}

Here, p(x) is the basis vector, and $\alpha$ contains the coefficients to be estimated. We use a quadratic polynomial basis in three dimensions:

\[
p(x)=[1,x,y,z,x^2,y^2,z^2,xy,yz,xz]^{\mathsf T}.
\]

\subsubsection{Coefficient Estimation}

The neighboring samples define the design matrix P and weight matrix W:

\begin{equation}
P=[p(x_1),p(x_2),\ldots,p(x_n)]^{\mathsf T}.
\label{eq:4}
\end{equation}

\begin{equation}
W=\operatorname{diag}\!\left(\omega(\lVert x-x_1\rVert),\omega(\lVert x-x_2\rVert),\ldots,\omega(\lVert x-x_n\rVert)\right).
\label{eq:5}
\end{equation}

The objective then takes the matrix form

\begin{equation}
\min_{\alpha}(y-P\alpha)^{\mathsf T}W(y-P\alpha).
\label{eq:6}
\end{equation}

With $y=[y_1,y_2,\ldots,y_n]^{\mathsf T}$, setting the derivative with respect to $\alpha$ to zero gives the coefficient vector $\alpha=(P^{\mathsf T}WP)^{-1}P^{\mathsf T}Wy$. The resulting coefficients define the local fitted surface. Projecting the original points onto this surface along its normal direction completes the filtering step \cite{ref6,ref32}.

\subsection{Three-Axis Reverse Completion}

To address holes in acquired point clouds, three-axis reverse completion follows a sequence of dimensionality reduction, interpolation within surface layers, reverse subdivision and clustering, multidirectional interpolation, and iterative refinement.

Dimensionality reduction projects the three-dimensional point cloud onto the planes normal to the X, Y, and Z axes, converting the problem into two-dimensional domains to reveal hole boundaries. Interpolation within each layer first improves point distribution before the remaining holes are processed. Reverse subdivision and clustering generate candidate hole points and cluster them to identify hole boundaries. Distance-weighted interpolation uses support from at least six directions to connect the surfaces interpolated across large holes with the surrounding point cloud boundaries and the overall surface. Iterative refinement then addresses persistent holes to complete the point cloud.

\subsubsection{Orthogonal Projection}

For the original point cloud $C=\{p(x,y,z)\}$, projection onto the $X=0$, $Y=0$, and $Z=0$ planes preserves the two coordinates within each plane and sets the perpendicular coordinate to zero. The projected points and their corresponding point sets are

On $Z=0$, $p_Z^{2D}=(x,y,0)$, with point set $C_Z^{2D}$.

On $X=0$, $p_X^{2D}=(0,y,z)$, with point set $C_X^{2D}$.

On $Y=0$, $p_Y^{2D}=(x,0,z)$, with point set $C_Y^{2D}$.

\subsubsection{Iterative Subdivision}

A two-dimensional concave hull defines the valid boundary of each projected point cloud. The hull vertices form the set B, and the extreme coordinates within the projection plane determine the subdivision domain. A coarse-to-fine subdivision strategy balances efficiency and resolution, while octree-based radius searches retrieve the sample points required for interpolation.

\subsubsection{Separation of Two Surface Layers}

Points from opposite surfaces may overlap when projected onto the same plane. For example, projecting the left and right walls of a room onto a plane normal to the left-to-right direction removes their separation in the projection, although their original perpendicular coordinates remain distinct. To distinguish the layers, we calculate the extrema $q_{\max}$ and $q_{\min}$ of the perpendicular coordinate q, determine whether the samples form one or two layers, and use gradient information to locate the layers. For two-layer regions, the point sets are

\begin{equation}
P_{\mathrm{up}}=\{P_i\mid q_i\in[q_{\max}-dth,q_{\max}]\}.
\label{eq:7}
\end{equation}

\begin{equation}
P_{\mathrm{down}}=\{P_i\mid q_i\in[q_{\min},q_{\min}+dth]\}.
\label{eq:8}
\end{equation}

Here, $dth$ is the layer thickness threshold.

\subsubsection{Multidirectional Inverse Distance Weighted Interpolation}

Interpolation uses observations from at least six directions to promote a smooth transition between new points and the surrounding surface. Both coordinates and colors use inverse squared distance weights, giving nearby samples greater influence:

\begin{equation}
w_i=\frac{1}{d_i^2}.
\label{eq:9}
\end{equation}

The distance $d_i=\sqrt{(x-x_i)^2+(y-y_i)^2}$ is measured within the projection plane, where x and y denote its two coordinates. The perpendicular coordinate is interpolated as

\begin{equation}
q=\frac{\sum_{i=1}^{m}w_iq_i}{\sum_{i=1}^{m}w_i}.
\label{eq:10}
\end{equation}

Here, q is the perpendicular coordinate, and m is the number of supporting samples. The same weighting applies to color. For the red channel r,

\begin{equation}
r=\frac{\sum_{i=1}^{m}w_ir_i}{\sum_{i=1}^{m}w_i}.
\label{eq:11}
\end{equation}

The green and blue channels follow the same rule. To reject outliers, we calculate the mean and mean absolute deviation of the perpendicular coordinates:

\begin{equation}
\mu=\frac{1}{m}\sum_{i=1}^{m}q_i.
\label{eq:12}
\end{equation}

\begin{equation}
\sigma=\frac{1}{m}\sum_{i=1}^{m}|q_i-\mu|.
\label{eq:13}
\end{equation}

Only samples satisfying $|q_i-\mu|\leq2\sigma$ contribute to interpolation.

\subsubsection{Reverse Hole Boundary Identification by Clustering}

The method identifies hole boundaries by clustering candidate points in missing regions. Density-based spatial clustering of applications with noise (DBSCAN) \cite{ref33} divides the candidate hole point cloud $C_{\mathrm{hole}}$ into separate regions $\{C_{\mathrm{hole},1},\ldots,C_{\mathrm{hole},K}\}$. For each region $C_{\mathrm{hole},i}$, concave hull reconstruction provides the boundary $B_i$, and a search in the original point cloud retrieves neighboring boundary samples $P_{w,i}$. Centered on the interpolation location, the projection plane is divided into six 60\textdegree{} sectors. The nearest point in each sector contributes to interpolation according to Section 3.3.4, providing support from multiple directions.

\subsubsection{Fusion of Completed Points}

The three-dimensional interpolated point sets $C_{\mathrm{ins},X}$, $C_{\mathrm{ins},Y}$, and $C_{\mathrm{ins},Z}$ from the three projection directions are combined into $C_{\mathrm{ins}}$. Merging this set with the original point cloud and removing duplicates yields $C_{\mathrm{full}}=C\cup C_{\mathrm{ins}}$.

\subsection{Comparison of Surface Reconstruction Algorithms}

We compare greedy projection triangulation \cite{ref24}, Marching Cubes \cite{ref27}, and Poisson surface reconstruction \cite{ref25} using the completed train nose point cloud.

\subsubsection{Greedy Projection Triangulation}

Greedy projection triangulation incrementally adds triangular faces according to local geometry to approximate the surface represented by the point cloud. The process defines local neighborhoods, constructs triangles using a greedy rule, and checks topological consistency.

For the input point cloud $P=\{p_1,p_2,\ldots,p_n\}$, we estimate normals and align their orientations with the local surface. A radius search or a k-nearest neighbors (k-NN) search defines the neighborhood of $p_i$:

\begin{equation}
\begin{aligned}N(p_i)&=\{p_j\mid\lVert p_j-p_i\rVert<r\}\quad\text{(radius neighborhood), or}\\N(p_i)&=\{p_j\mid\operatorname{rank}(\lVert p_j-p_i\rVert)\leq k\}\quad\text{(k-NN neighborhood).}\end{aligned}
\label{eq:14}
\end{equation}

Here, r is the search radius, and k is the number of neighbors. Starting from a seed point, the algorithm follows a shortest-edge-first greedy rule when connecting neighboring points to form a triangular face $\Delta p_i p_j p_k$. The angle between the face normal and the local point normals must remain within the threshold $\theta$, and each new face undergoes a topological consistency check against the existing mesh. Repeating this process produces a connected triangular surface.

\subsubsection{Marching Cubes}

Marching Cubes (MC) discretizes three-dimensional space into a regular voxel grid and reconstructs a surface from intersections between voxel edges and a target isosurface. The procedure requires a scalar field and interpolation of the edge intersections.

A regular three-dimensional voxel grid covers the spatial extent of the point cloud. At each of the eight vertices $v=(x,y,z)$ of a voxel, the signed distance is

\begin{equation}
f(v)=\operatorname{sign}(v)\cdot\min_{p\in P}\lVert v-p\rVert.
\label{eq:15}
\end{equation}

The term $\operatorname{sign}(v)$ indicates whether a vertex lies inside or outside the surface, and P denotes the input point cloud. Let T be the isovalue. For $T=0$, a voxel contains part of the isosurface if two vertices $i\ne j$ satisfy $f(v_i)f(v_j)<0$. For a general T, the condition becomes $[f(v_i)-T][f(v_j)-T]<0$. Linear interpolation along an intersected edge gives the point q satisfying $f(q)=T$:

\begin{equation}
q=v_a+\frac{T-f(v_a)}{f(v_b)-f(v_a)}(v_b-v_a).
\label{eq:16}
\end{equation}

Here, $v_a$ and $v_b$ are the endpoints of the voxel edge. A predefined connectivity template links the intersections into triangles, and the triangles from adjacent voxels form the reconstructed surface.

\subsubsection{Poisson Surface Reconstruction}

Poisson surface reconstruction builds a global scalar field by solving a Poisson equation. Point normals provide gradient constraints, and an isosurface of the resulting field defines the reconstructed surface.

Normal estimation and consistent orientation of the input point cloud P produce the oriented samples $\{(p_i,n_i)\}$. Trilinear interpolation distributes the sample normals to neighboring octree nodes, with weights that account for local sampling density. Locally supported basis functions at these nodes define the vector field $V:\mathbb R^3\to\mathbb R^3$ \cite{ref25}. Reconstruction then seeks an implicit function $\varphi$ whose gradient approximates V:

\begin{equation}
\min_{\varphi}\int_{\mathbb R^3}\lVert\nabla\varphi(x)-V(x)\rVert_2^2\,dx.
\label{eq:17}
\end{equation}

The corresponding Euler-Lagrange equation gives the Poisson equation:

\begin{equation}
\Delta\varphi(x)=\nabla\cdot V(x).
\label{eq:18}
\end{equation}

The operator $\Delta=\partial^2/\partial x^2+\partial^2/\partial y^2+\partial^2/\partial z^2$ is the Laplacian, and $\nabla\cdot V$ is the divergence of V. Discretization on an adaptive octree converts the equation into the linear system

\begin{equation}
A\varphi=b.
\label{eq:19}
\end{equation}

Here, A is the discrete Laplacian matrix, b is the discretized divergence term, and $\varphi$ in Equation (19) is the coefficient vector of the implicit function in the locally supported basis. A conjugate gradient (CG) solver computes the coefficients. The isovalue follows from the function values at the input samples, $\gamma=\frac{\sum_i a_i\varphi(p_i)}{\sum_i a_i}$, where $a_i$ is inversely related to local sampling density; uniform sampling permits equal weights. An octree-compatible Marching Cubes procedure extracts the isosurface $\varphi=\gamma$ as the initial mesh \cite{ref25}. Mesh simplification, smoothing, and topology refinement then improve the surface quality.

\section{Experimental Results and Analysis}

\subsection{Point Cloud Acquisition Results}

The experiments use a proportionally scaled display model of a high-speed train nose. Its streamlined geometry includes cab windows, flow-guiding grooves, and lower skirt structures. The research team acquires the data through handheld mobile scanning with the custom multisensor device.

\subsubsection{Acquired Data}

FAST-LIVO2 registers the acquired multisensor data \cite{ref5} to produce an initial point cloud of 9,706,530 points. The cloud covers the main exterior of the display model, while local gaps remain around the windows, as shown in Figure 2.

\begin{figure}[htbp]
\centering
\includegraphics[width=\linewidth]{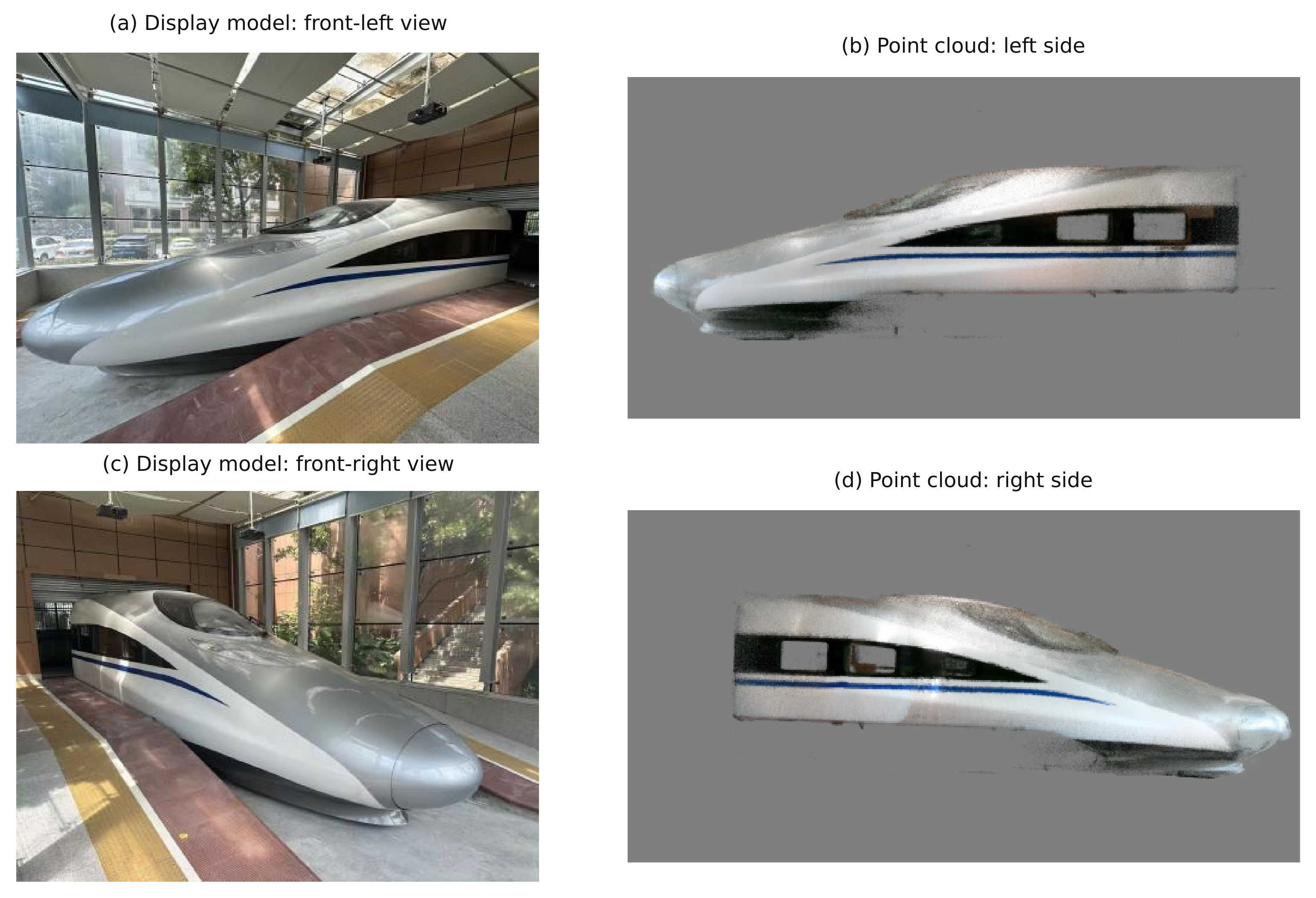}
\caption{The proportionally scaled high-speed train nose display model and its scanned point cloud. Panels (a) and (c) show photographs from different viewing directions; panels (b) and (d) show the corresponding side views of the point cloud.}
\label{fig:2}
\end{figure}

\subsection{Comparison of Filtering Results}

We compare random downsampling with MLS smoothing, as shown in Figure 3. MLS combines local surface fitting with smooth sampling \cite{ref6,ref32}. Table 1 reports the results for the acquired display model point cloud.

\begin{figure}[htbp]
\centering
\includegraphics[width=\linewidth]{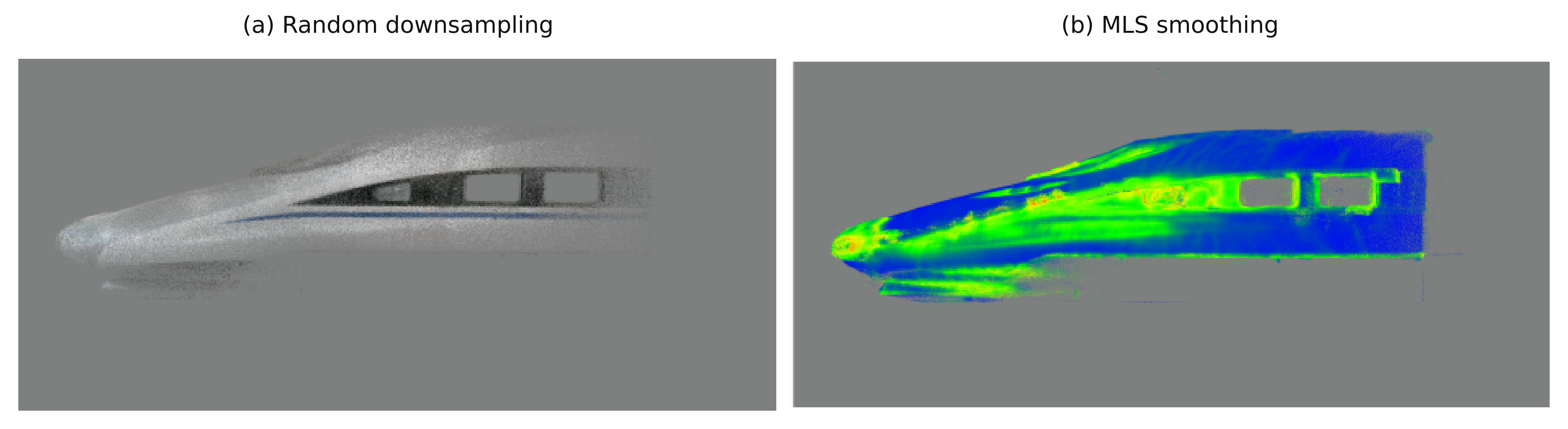}
\caption{Results of random downsampling and MLS smoothing. Each panel shows the processed point cloud using the original display colors. Table 1 provides the quantitative comparison.}
\label{fig:3}
\end{figure}

\begin{table}[htbp]
\centering
\caption{Comparison of filtering results.}
\label{tab:1}
\scriptsize
\setlength{\tabcolsep}{3pt}
\renewcommand{\arraystretch}{1.3}
\begin{tabularx}{\linewidth}{@{}YYYYYYYY@{}}
\toprule

Method & Input points & Output points & Point count reduction (\%) & Density change (\%) & Uniformity change & Hausdorff distance (cm) & Chamfer distance (cm) \\

\midrule

MLS smoothing & 9706530 & 1118817 & 88.473564 & 88.470909 & 5.454283 & 1.7564 & 0.6927 \\

Random downsampling & 9706530 & 970653 & 90.000000 & 89.575424 & 2.464060 & 99.1275 & 0.7447 \\

\bottomrule
\end{tabularx}
\end{table}

\subsubsection{Analysis}

Table 1 shows that MLS reduces the point count from 9,706,530 to 1,118,817, a reduction of 88.47\%. The density change is also 88.47\%, consistent with the change in point count. The uniformity change metric is 5.454 for MLS and 2.464 for random downsampling, indicating a larger improvement in point distribution uniformity for MLS in this experiment.

MLS yields a Hausdorff distance of 1.7564 cm and a Chamfer distance of 0.6927 cm, both lower than those of random downsampling. The difference is particularly large for the Hausdorff distance, which is 99.1275 cm for random downsampling. These results indicate that MLS reduces the number of points while preserving the spatial distribution and geometry more closely under the evaluation used here.

\subsection{Analysis of Sparse Point Cloud Densification by Interpolation}

We compare three-axis reverse completion with nearest-neighbor interpolation. Figure 4 shows the three orthogonal projection directions, and Figure 5 presents top and side views before and after completion. The proposed method fills missing points around the windows. Some gaps remain near the rear of the display model because boundary support is insufficient.

\begin{figure}[htbp]
\centering
\includegraphics[width=\linewidth]{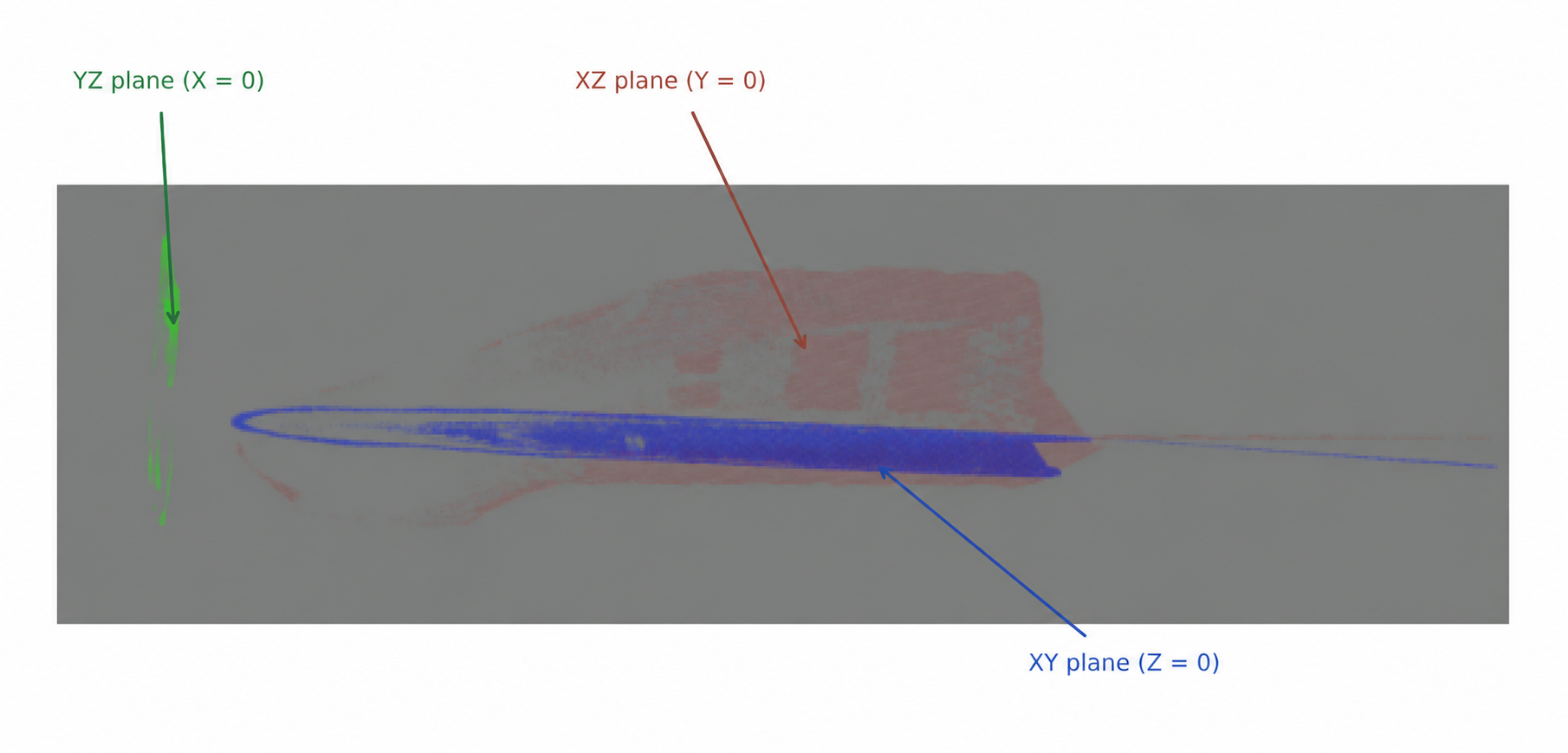}
\caption{Spatial arrangement of the three orthogonal projected point sets. Green, red, and blue indicate the YZ, XZ, and XY projection planes, respectively, following the coordinate plane definitions in Section 3.3.1.}
\label{fig:4}
\end{figure}

\begin{figure}[htbp]
\centering
\includegraphics[width=\linewidth]{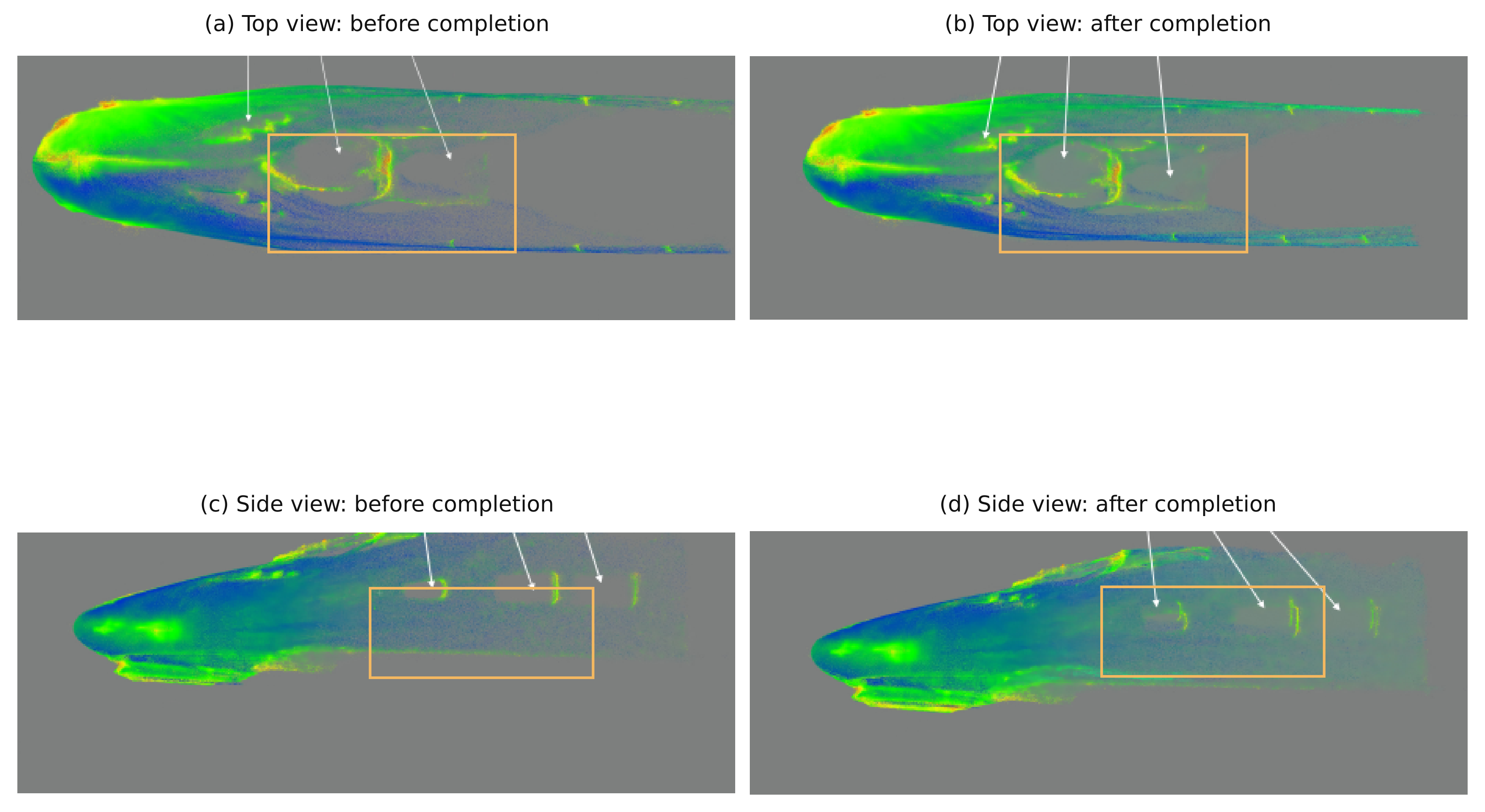}
\caption{Point clouds before and after three-axis reverse completion. Panels (a) and (b) show top views, and panels (c) and (d) show side views. The left and right columns show the results before and after completion, respectively. Orange boxes mark the window regions used for comparison.}
\label{fig:5}
\end{figure}

\begin{table}[htbp]
\centering
\caption{Comparison of point cloud completion results. Three-axis reverse completion denotes the complete procedure in Section 3.3.}
\label{tab:2}
\footnotesize
\setlength{\tabcolsep}{3pt}
\renewcommand{\arraystretch}{1.3}
\begin{tabularx}{\linewidth}{@{}YYYY@{}}
\toprule

Method & Point count & Coefficient of determination $R^2$ & RMSE (cm) \\

\midrule

Three-axis reverse completion & 7,679,247 & 1 & 0.118 \\

Nearest-neighbor interpolation & 7,994,022 & 0.9879 & 146.907 \\

\bottomrule
\end{tabularx}
\end{table}

\subsubsection{Analysis}

Table 2 compares the completion results. Three-axis reverse completion increases the point count to 7,679,247, with a coefficient of determination $R^2$ of 1 and a root mean square error (RMSE) of approximately 0.118 cm. Nearest-neighbor interpolation produces 7,994,022 points, with $R^2$ = 0.9879 and RMSE = 146.907 cm.

\subsection{Comparison of Surface Reconstruction Results}

Greedy projection triangulation \cite{ref24}, Poisson surface reconstruction \cite{ref25}, and the MC-based pipeline \cite{ref27} use the completed point cloud as input. Figure 6 shows their outputs. In this experiment, greedy projection triangulation follows the input point cloud contour more closely, whereas the other two pipelines show deviations in the overall reconstructed shape.

\begin{figure}[htbp]
\centering
\includegraphics[width=\linewidth]{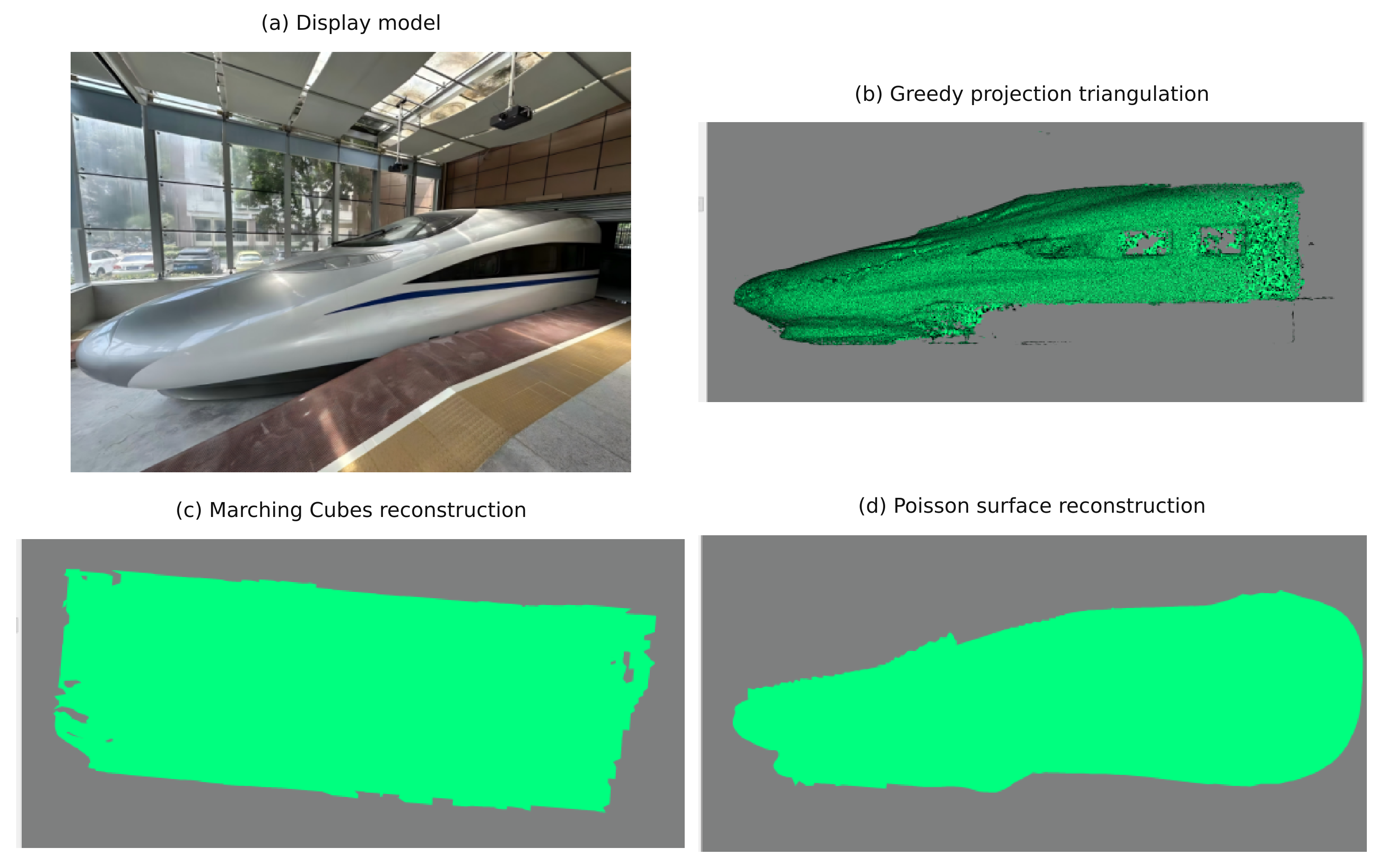}
\caption{Outputs of the three surface reconstruction pipelines. Panel (a) shows the display model; panels (b), (c), and (d) show greedy projection triangulation, the MC-based pipeline, and Poisson surface reconstruction, respectively.}
\label{fig:6}
\end{figure}

\begin{table}[htbp]
\centering
\caption{Comparison of surface reconstruction results.}
\label{tab:3}
\scriptsize
\setlength{\tabcolsep}{3pt}
\renewcommand{\arraystretch}{1.3}
\begin{tabularx}{\linewidth}{@{}YYYYYYYY@{}}
\toprule

Method & Hausdorff distance & Maximum deviation distance & Mean deviation distance & Standard deviation & RMSE & Surface smoothness & Face count \\

\midrule

Greedy projection triangulation & 0.0086 cm & 0.0086 cm & 0.0036 cm & 0.0016 cm & 0.0040 cm & 0.876081 & 12182654 \\

Poisson surface reconstruction & 53.0692 cm & 54.1552 cm & 2.2900 cm & 1.4693 cm & 2.7209 cm & 0.873940 & 111600 \\

MC-based reconstruction & 481.0731 cm & 46.2680 cm & 6.3177 cm & 3.1728 cm & 7.0697 cm & 0.000000 & 50597 \\

\bottomrule
\end{tabularx}
\end{table}

\subsubsection{Analysis}

Table 3 summarizes the evaluation of the three reconstruction pipelines.

Greedy projection triangulation yields a Hausdorff distance of 0.0086 cm, compared with 53.0692 cm for Poisson reconstruction and 481.0731 cm for the MC-based pipeline. The maximum deviation distances are 0.0086 cm, 54.1552 cm, and 46.2680 cm, respectively. Greedy projection triangulation also has the lowest mean deviation, standard deviation, and RMSE, at 0.0036 cm, 0.0016 cm, and 0.0040 cm. It therefore produces the lowest geometric distance errors among the three pipelines under the evaluation setting used in this study.

The surface smoothness metrics of greedy projection triangulation and Poisson reconstruction are similar, at 0.876081 and 0.873940, respectively. The MC-based pipeline has a value of 0 for this metric. Greedy projection triangulation generates 12,182,654 faces, more than either alternative, providing a finer subdivision of local surfaces.

Considering geometric distance errors, surface smoothness, and contour preservation in this experiment, we use greedy projection triangulation to reconstruct the surface from the completed point cloud.

\section{Conclusion}

This study presents a pipeline that combines FAST-LIVO2 multisensor mapping, MLS smoothing, three-axis reverse completion, and surface reconstruction to address sparse sampling and missing glass window regions in train nose scans. Experiments use a proportionally scaled display model of a high-speed train nose. The main conclusions are as follows:

\begin{enumerate}

\item MLS smoothing reduces the point count while preserving the geometric features of the train nose. The point count decreases by 88.47\%, the Hausdorff and Chamfer distances are lower than those of random downsampling, and the point distribution becomes more uniform.

\item The proposed three-axis projection-based subdivision and interpolation with reverse hole boundary identification fills missing window regions through multidirectional projection, hole localization, and interpolation. Under the evaluation setting used in this study, the completed point cloud has an RMSE of 0.118 cm, lower than that of nearest-neighbor interpolation.

\item Among the three surface reconstruction algorithms, greedy projection triangulation preserves the train nose contour and curvature features well. Its Hausdorff distance of 0.0086 cm is lower than those of Poisson surface reconstruction and the Marching Cubes-based pipeline, making it suitable for reconstructing surfaces from the completed point cloud in this study.

\end{enumerate}

By linking sparse point cloud completion with surface reconstruction, this study establishes a pipeline for non-contact digital modeling of a train nose display model. Three-axis reverse completion uses the geometry around holes to identify and fill missing regions, improving data completeness near the glass windows and providing more continuous point cloud input for subsequent surface reconstruction. This workflow offers a practical basis for representing and analyzing train nose geometry.

This study primarily validates the pipeline on a proportionally scaled display model and does not yet cover the range of acquisition conditions found in complex industrial environments. Three-axis reverse completion depends on valid boundary information around each hole. Insufficient boundary support leaves some regions near the rear of the model incomplete.

\FloatBarrier


\begin{thebibliography}{99}

\bibitem{ref1} LI M, LIU N, WANG D H. 2019. Design and aerodynamic performance of a new nose shape for a 500 km/h high-speed train. Journal of Mechanical Engineering, 55(18): 142--149 (in Chinese). \url{https://doi.org/10.3901/JME.2019.18.142}

\bibitem{ref2} WANG S, YOU P, WANG H, et al. 2024. PDE parametric modeling with a two-stage MLP for aerodynamic shape optimization of high-speed train heads. Structural and Multidisciplinary Optimization, 67: 169. \url{https://doi.org/10.1007/s00158-024-03886-9}

\bibitem{ref3} YUN S, KWAK M, KANG H, et al. 2024. Two-stage multi-objective optimization for improving the aerodynamic characteristics of high-speed train nose shape. Fluids, 9(12): 295. \url{https://doi.org/10.3390/fluids9120295}

\bibitem{ref4} XU W, CAI Y, HE D, et al. 2022. FAST-LIO2: Fast direct LiDAR-inertial odometry. IEEE Transactions on Robotics, 38(4): 2053--2073. \url{https://doi.org/10.1109/TRO.2022.3141876}

\bibitem{ref5} ZHENG C, XU W, ZOU Z, et al. 2025. FAST-LIVO2: Fast, direct LiDAR-inertial-visual odometry. IEEE Transactions on Robotics, 41: 326--346. \url{https://doi.org/10.1109/TRO.2024.3502198}

\bibitem{ref6} ALEXA M, BEHR J, COHEN-OR D, et al. 2003. Computing and rendering point set surfaces. IEEE Transactions on Visualization and Computer Graphics, 9(1): 3--15. \url{https://doi.org/10.1109/TVCG.2003.1175093}

\bibitem{ref7} YU L, LI X, FU C W, et al. 2018. PU-Net: Point cloud upsampling network. Proceedings of the IEEE/CVF Conference on Computer Vision and Pattern Recognition: 2790--2799. \url{https://openaccess.thecvf.com/content_cvpr_2018/html/Yu_PU-Net_Point_Cloud_CVPR_2018_paper.html}

\bibitem{ref8} QIAN G, ABUALSHOUR A, LI G, et al. 2021. PU-GCN: Point cloud upsampling using graph convolutional networks. Proceedings of the IEEE/CVF Conference on Computer Vision and Pattern Recognition: 11683--11692. \url{https://openaccess.thecvf.com/content/CVPR2021/html/Qian_PU-GCN_Point_Cloud_Upsampling_Using_Graph_Convolutional_Networks_CVPR_2021_paper.html}

\bibitem{ref9} TESEMA K W, HILL L, JONES M W, et al. 2024. Point cloud completion: A survey. IEEE Transactions on Visualization and Computer Graphics, 30(10): 6880--6899. \url{https://doi.org/10.1109/TVCG.2023.3344935}

\bibitem{ref10} WANG J, CUI Y, GUO D, et al. 2024. PointAttN: You only need attention for point cloud completion. Proceedings of the AAAI Conference on Artificial Intelligence, 38(6): 5472--5480. \url{https://doi.org/10.1609/aaai.v38i6.28356}

\bibitem{ref11} DU Y, ZHAO Z, SU S, et al. 2025. SuperPC: A single diffusion model for point cloud completion, upsampling, denoising, and colorization. Proceedings of the IEEE/CVF Conference on Computer Vision and Pattern Recognition: 16953--16964. \url{https://openaccess.thecvf.com/content/CVPR2025/html/Du_SuperPC_A_Single_Diffusion_Model_for_Point_Cloud_Completion_Upsampling_CVPR_2025_paper.html}

\bibitem{ref12} UDDIN K, JEONG T H, OH B T. 2022. Incomplete region estimation and restoration of 3D point cloud human face datasets. Sensors, 22(3): 723. \url{https://doi.org/10.3390/s22030723}

\bibitem{ref13} GOU G, SUI H, LI D, et al. 2022. LIMOFilling: Local information guide hole-filling and sharp feature recovery for manifold meshes. Remote Sensing, 14(2): 289. \url{https://doi.org/10.3390/rs14020289}

\bibitem{ref14} YIP M, STAHL A, SCHELLEWALD C. 2024. Robust hole-detection in triangular meshes irrespective of the presence of singular vertices. Computer-Aided Design, 170: 103696. \url{https://doi.org/10.1016/j.cad.2024.103696}

\bibitem{ref15} KANG S, JI L Y, JIAO Q S, et al. 2020. Comparison of interpolation methods for terrestrial LiDAR point cloud data. Journal of Geodesy and Geodynamics, 40(4): 400--404 (in Chinese). \url{https://doi.org/10.14075/j.jgg.2020.04.015}

\bibitem{ref16} LAN Y L, KANG C L, WANG N, et al. 2023. Point cloud hole filling using moving least squares with additional increment conditions. Infrared and Laser Engineering, 52(2): 20220390 (in Chinese). \url{https://doi.org/10.3788/IRLA20220390}

\bibitem{ref17} ZHANG C, ZHOU H, DUAN J. 2023. A method for identifying and repairing holes on the surface of unorganized point cloud. Measurement, 210: 112575. \url{https://doi.org/10.1016/j.measurement.2023.112575}

\bibitem{ref18} HOU Y Q, WANG C Y, HUAI S R, et al. 2025. Multiple-hole filling for unorganized point cloud models. Journal of Zhejiang University (Engineering Science), 59(10): 2045--2055 (in Chinese). \url{https://doi.org/10.3785/j.issn.1008-973X.2025.10.005}

\bibitem{ref19} XIE Y, XIE X, WANG Y, et al. 2024. Terrain point cloud inpainting via signal decomposition. Computers \& Graphics, 120: 103915. \url{https://doi.org/10.1016/j.cag.2024.103915}

\bibitem{ref20} SUN C, MIAO L, WANG M, et al. 2023. Research on point cloud hole filling and 3D reconstruction in reflective area. Scientific Reports, 13: 18524. \url{https://doi.org/10.1038/s41598-023-45648-5}

\bibitem{ref21} LI R, LI X, FU C W, et al. 2019. PU-GAN: A point cloud upsampling adversarial network. Proceedings of the IEEE/CVF International Conference on Computer Vision: 7202--7211. \url{https://doi.org/10.1109/ICCV.2019.00730}

\bibitem{ref22} DU H, YAN X, WANG J, et al. 2024. Arbitrary-scale point cloud upsampling by voxel-based network with latent geometric-consistent learning. Proceedings of the AAAI Conference on Artificial Intelligence, 38(2): 1626--1634. \url{https://doi.org/10.1609/aaai.v38i2.27929}

\bibitem{ref23} KHATER M, STRAUSS M, VON OLSHAUSEN P, et al. 2025. PU-Gaussian: Point cloud upsampling using 3D Gaussian representation. Proceedings of the IEEE/CVF International Conference on Computer Vision Workshops: 2644--2652. \url{https://doi.org/10.1109/ICCVW69036.2025.00275}

\bibitem{ref24} MARTON Z C, RUSU R B, BEETZ M. 2009. On fast surface reconstruction methods for large and noisy point clouds. Proceedings of the IEEE International Conference on Robotics and Automation: 3218--3223. \url{https://doi.org/10.1109/ROBOT.2009.5152628}

\bibitem{ref25} KAZHDAN M, BOLITHO M, HOPPE H. 2006. Poisson surface reconstruction. Proceedings of the Symposium on Geometry Processing: 61--70. \url{https://doi.org/10.2312/SGP/SGP06/061-070}

\bibitem{ref26} KAZHDAN M, HOPPE H. 2013. Screened Poisson surface reconstruction. ACM Transactions on Graphics, 32(3): 29. \url{https://hhoppe.com/proj/screenedpoisson/}

\bibitem{ref27} LORENSEN W E, CLINE H E. 1987. Marching cubes: A high resolution 3D surface construction algorithm. ACM SIGGRAPH Computer Graphics, 21(4): 163--169. \url{https://doi.org/10.1145/37402.37422}

\bibitem{ref28} HUANG J, GOJCIC Z, ATZMON M, et al. 2023. Neural kernel surface reconstruction. Proceedings of the IEEE/CVF Conference on Computer Vision and Pattern Recognition: 4369--4379. \url{https://openaccess.thecvf.com/content/CVPR2023/html/Huang_Neural_Kernel_Surface_Reconstruction_CVPR_2023_paper.html}

\bibitem{ref29} SHI C, TANG F, AN N, et al. 2025. 3D-SLNR: A super lightweight neural representation for large-scale 3D mapping. Proceedings of the IEEE/CVF Conference on Computer Vision and Pattern Recognition: 27233--27242. \url{https://openaccess.thecvf.com/content/CVPR2025/html/Shi_3D-SLNR_A_Super_Lightweight_Neural_Representation_for_Large-scale_3D_Mapping_CVPR_2025_paper.html}

\bibitem{ref30} XIE Y, HUANG Z, WU J, et al. 2025. GS-LIVM: Real-time photo-realistic LiDAR-inertial-visual mapping with Gaussian splatting. Proceedings of the IEEE/CVF International Conference on Computer Vision: 26869--26878. \url{https://openaccess.thecvf.com/content/ICCV2025/html/Xie_GS-LIVM_Real-Time_Photo-Realistic_LiDAR-Inertial-Visual_Mapping_with_Gaussian_Splatting_ICCV_2025_paper.html}

\bibitem{ref31} MACENSKI S, FOOTE T, GERKEY B, et al. 2022. Robot Operating System 2: Design, architecture, and uses in the wild. Science Robotics, 7(66): eabm6074. \url{https://doi.org/10.1126/scirobotics.abm6074}

\bibitem{ref32} KANG C L, LU T N, ZONG M M, et al. 2020. Point cloud smooth sampling and surface reconstruction based on moving least squares. The International Archives of the Photogrammetry, Remote Sensing and Spatial Information Sciences, XLII-3/W10: 145--151. \url{https://doi.org/10.5194/isprs-archives-XLII-3-W10-145-2020}

\bibitem{ref33} ESTER M, KRIEGEL H P, SANDER J, et al. 1996. A density-based algorithm for discovering clusters in large spatial databases with noise. Proceedings of the Second International Conference on Knowledge Discovery and Data Mining: 226--231.

\end{thebibliography}
\end{document}